# Predicting person-level injury severity using crash narratives: A balanced approach with roadway classification and natural language process techniques

Mohammad Zana Majidi[1*], Sajjad Karimi[2], Teng Wang[3], Robert Kluger[2], Reginald Souleyrette[1]

[1] Department of Civil Engineering, University of Kentucky, KY, USA. Email: zana.majidi@uky.edu; souleyrette@uky.edu.
[2] Department of Civil and Environmental Engineering, University of Louisville, KY, USA. Email: sajjad.karimizinkanlu@louisville.edu; robert.kluger@louisville.edu.
[3] Kentucky Transportation Center, Lexington, KY, USA. Email: wangtengyilang@gmail.com.

## Abstract

Predicting injuries and fatalities in traffic crashes plays a critical role in enhancing road safety, improving emergency response, and guiding public health interventions. This study investigates the added value of unstructured crash narratives (written by police officers at the scene) when combined with structured crash data to predict injury severity. Two widely used Natural Language Processing (NLP) techniques, Term Frequency–Inverse Document Frequency (TF-IDF) and Word2Vec, were employed to extract semantic meaning from the narratives, and their effectiveness was compared. To address the challenge of class imbalance, a K-Nearest Neighbors–based oversampling method was applied to the training data prior to modeling. The dataset consists of crash records from Kentucky spanning 2019 to 2023. To account for roadway heterogeneity, three road classification schemes were used: (1) eight detailed functional classes (e.g., Urban Two-Lane, Rural Interstate, Urban Multilane Divided), (2) four broader paired categories (e.g., Urban vs. Rural, Freeway vs. Non-Freeway), and (3) a unified dataset without classification. A total of 102 machine learning models were developed by combining structured features and narrative-based features using the two NLP techniques alongside three ensemble algorithms: XGBoost, Random Forest, and AdaBoost. Results demonstrate that models incorporating narrative data consistently outperform those relying solely on structured data. Among all combinations, TF-IDF coupled with XGBoost yielded the most accurate predictions in most subgroups. The findings highlight the power of integrating textual and structured crash information to enhance person-level injury prediction. This work offers a practical and adaptable framework for transportation safety professionals to improve crash severity modeling, guide policy decisions, and design more effective countermeasures.

## Keywords

Individual injury prediction, Crash severity, Narrative data, Natural Language Processing, TF-IDF, Word2Vec, Machine learning, Road classification, Oversampling.

## 1. Introduction

Accurate prediction of injury severity in traffic crashes is a fundamental concern in transportation safety and public health, as it informs emergency response planning, medical resource allocation, and the development of targeted safety interventions. An ongoing challenge in the Kentucky crash dataset is that over 50% of individuals involved in crashes during the 2019–2023 period have either missing or unreported injury severity information. While most existing studies rely solely on structured crash data such as vehicle type, roadway geometry, or weather conditions, an often overlooked yet rich source of information lies in the narrative descriptions written by police officers at the crash scene. These unstructured narratives frequently contain key contextual insights that are not captured by structured fields, such as pre-crash vehicle movements, contributing behaviors, or sequence of events.

This study investigates the added value of unstructured crash narratives when combined with structured crash data to improve injury severity prediction, focusing on the individual level. The objective is to apply this hybrid approach, using two established NLP methods, across multiple road classification categories and real-world scenarios to improve both the precision and generalizability of the predictions. Two well-established Natural Language Processing (NLP) methods, TF-IDF and Word2Vec, are employed to transform narrative text into numerical vectors, allowing their integration into machine learning models. TF-IDF is a frequency-based technique that helps identify the most informative terms by evaluating their importance within and across documents (*1*). Word2Vec transforms words into dense vector representations that preserve semantic and contextual meaning (*2*). To address the issue of class imbalance in injury outcomes, a K-Nearest Neighbors–based oversampling strategy is applied to a training portion of the data before modeling. Moreover, to ensure flexibility and capture roadway-specific effects, the dataset is stratified using three different road classification schemes: (1) a detailed classification with eight road types (e.g., Urban Two-Lane, Freeway, etc.), (2) a broader paired grouping based on geometric or functional similarities (e.g., Urban vs. Rural, Divided vs. Undivided), and (3) an undivided dataset representing the overall crash population. This comprehensive design allows for evaluating how the combination of narrative text, structured data, and road classification affects model performance, offering a more nuanced and scalable approach to injury severity prediction. In this study, injury prediction models are first developed using only structured crash data across the three road classification groups. Following this, narrative crash descriptions are incorporated to assess their added value in improving predictive performance. All models are trained using XGBoost, RF, and AdaBoost across the three road classification groups. The comparison of results across these configurations clearly shows that adding narrative information improves model performance in almost all cases. Prior studies have shown that combining narrative text with structured data can significantly improve model performance in crash analysis (*3*). The results of this study support these findings, showing that models integrating narrative information achieved higher accuracy and better overall performance than those relying on structured data alone.

By applying a dependable prediction framework capable of estimating unknown values with high accuracy, this study addresses a gap and provides more robust information for evidence-based policy decisions. These person-level predictions can support transportation safety agencies and decision-makers in crafting more targeted and effective measures aimed at lowering both the occurrence and severity of crash-related injuries and fatalities. Additionally, by analyzing the sensitivity of key input features, this approach can offer valuable insights into how regulatory changes, such as adjustments to speed limits, might influence injury outcomes, aiding in more informed planning and policy evaluation.

The remainder of this paper is organized as follows: Section 2 provides a review of relevant literature on using narratives and related techniques for crash and injury severity prediction. Section 3 describes the dataset, preprocessing steps, and road classification methodology. Section 4 outlines the machine learning models, NLP approaches used, and the oversampling technique applied to address data imbalance. Section 5 presents the results, model comparisons, and interpretation of findings. Finally, Section 6 concludes the study and offers recommendations for future research and safety policy.

# 2. Literature Review

Recent research has broadly classified crash severity prediction approaches into three major categories: probabilistic modeling techniques, machine learning algorithms, and natural language processing (NLP) methods (*4*).

Injury severity prediction in traffic crashes has traditionally been approached as a classification task, where outcomes are divided into distinct severity levels such as fatal, severe, minor, or property-damage-only crashes. One of the most established methodologies in this domain involves probabilistic models, which formulate crash severity prediction as a function of conditional probabilities based on input features. These models often adopt a regression-based framework, notably logistic or ordered logit/probit regressions, in which the dependent variable corresponds to a severity category (*4*).

For example, Fu et al. applied a bivariate generalized ordered probability model to jointly model the severity outcomes for both directions involved in a collision, effectively capturing intra-collision correlation and asymmetry in severity across risk types (*5*). This multivariate probabilistic approach enabled a more refined classification of crash severity based on mutual influences between parties involved. Similarly, Christoforou et al. employed a stochastic parameter-ordered probability model to evaluate how traffic volume and vehicle speed contribute to injury outcomes, showing that stochastic variations are particularly significant in high-density traffic environments (*6*). Other studies have employed ordered probit or logit models to explore the influence of factors such as vehicle type, (*7-9*) .

Further contributions have addressed vulnerable road users. For instance, Do et al. implemented a stochastic parametric negative binomial (RPNB) framework nested within a binary logit architecture to quantify the impact of location-based and crash-specific variables on injury outcomes for cyclists and pedestrians (*10*). These models provide insights into severity categorization using more flexible distributions, especially in data with overdispersion or zero-inflated values. Moreover, other studies, such as “Factors Associated with Vulnerable Road Users on Freeways in North Carolina”, have focused on identifying factors associated with these vulnerable users (*11*). While these probabilistic models are interpretable and statistically sound, they often rely on strong assumptions like variable independence or linearity, which can limit their applicability in complex and nonlinear datasets.

While probabilistic models offer interpretability, their performance can suffer when modeling complex nonlinear relationships or interacting variables. This has led to the increasing adoption of machine learning (ML) models for traffic crash severity prediction. These models offer the flexibility to capture high-dimensional feature interactions without relying on strict statistical assumptions. Among ML traditional techniques, decision tree models have been widely applied for their intuitive structure and ability to handle both categorical and numerical features. Süleyman and Albayrak. used a decision tree to evaluate the contribution of driving behavior and demographic parameters on cyclist injury severity, while also employing a fuzzy logic system to improve classification accuracy under uncertainty (*12*).

To address these limitations, ensemble models such as Random Forest (RF) and Extreme Gradient Boosting (XGBoost) have become popular. For instance, Ruangkanjanases et al. showed that artificial neural networks (ANNs) could significantly improve crash prediction accuracy compared to decision trees, particularly when modeling continuous injury scores (*13*). More advanced models like deep neural networks (DNNs) and convolutional neural networks (CNNs) have also been explored for their superior pattern recognition capabilities. Jamal et al. reported that XGBoost outperformed both RF and traditional decision trees when predicting crash damage severity, thanks to its regularization mechanisms and scalability (*14*). A particularly notable development came from Niyogisubizo et al., who proposed a hybrid stacking framework that integrated XGBoost, AdaBoost, and a multilayer perception (MLP) to leverage complementary strengths of each model. Their ensemble strategy achieved substantial gains in both injury and property-damage-only severity predictions (*15*).

To mitigate data imbalance, a common problem in crash datasets, Cai et al. introduced a deep convolutional generative adversarial network (DCGAN) to synthetically augment rare crash events, which significantly enhanced model robustness (*16*). Rahim et al. applied CNNs in combination with temporal analysis, demonstrating superior accuracy over classical methods like SVM, especially for predicting fatal crashes (*17*). Islam et al. further improved upon these approaches by integrating deep learning with real-time data streams, allowing for responsive prediction systems capable of handling live crash data feeds (*18*).

Traditional crash severity prediction relies almost exclusively on structured datasets, coded fields such as vehicle type, weather, road condition, and crash configuration. However, these fields often fail to capture the full context of an event. Narrative data, typically recorded by police officers at crash scenes, include rich qualitative descriptions that can reveal crucial behavioral or environmental factors not represented in structured variables. The emergence of Natural Language Processing (NLP) offers new avenues for leveraging these unstructured narratives. Fan et al. pioneered the idea of converting structured crash records into textual format to enable textual reasoning using Large Language Models (LLMs) for predicting crash severity. By framing structured data as natural language, they demonstrated that LLMs could capture latent patterns that standard algorithms often miss (*19*). Li et al. took a different approach by applying a Structural Topic Model (STM) combined with XGBoost to identify latent topics in narrative crash reports and relate them to severity outcomes. This method proved especially useful in uncovering underlying themes, such as distraction, aggression, or hazardous road design, that are typically not encoded in traditional variables (*20*).

Beyond these, established NLP techniques like Term Frequency-Inverse Document Frequency (TF-IDF) and Word2Vec (W2V) have shown utility in quantifying narrative text. TF-IDF evaluates the importance of words across crash reports, helping to isolate significant terms that correlate with injury severity. Meanwhile, Word2Vec, which transforms words into dense vector embeddings based on context, enables the model to understand semantic relationships in narratives. When combined with ensemble models such as XGBoost or RF, both TF-IDF and W2V representations have improved classification performance over models trained on structured data alone (*1; 21; 22*).

Additionally, transformer-based models, such as BERT and GPT variants, are increasingly being employed for end-to-end severity prediction. These models, as demonstrated by Jiang et al. (*6*), excel at processing long-form narratives and identifying abstract concepts like driver intent or evasive maneuvers. Such capabilities allow for deeper contextual understanding, significantly enhancing the predictive accuracy of severity models that include unstructured inputs. NLP-driven frameworks thus represent a promising direction in crash severity modeling, particularly in contexts where structured data is incomplete or lacks nuance. By extracting and interpreting narrative insights, these models can offer a more comprehensive and accurate picture of crash risk.

# 3. Available Data

Accurate data collection and meticulous preparation are fundamental to the integrity of any research study. The use of erroneous values, inconsistent records, or poorly integrated datasets can compromise the validity and reliability of results. In this study, considerable effort was devoted to the data preparation and preprocessing phase due to the involvement of multiple sources and heterogeneous data types. As illustrated in Figure 1, the primary data for this research was obtained from two key sources: the Kentucky Highway Information System (HIS) and the official crash records maintained by the Kentucky State Police.

## 3.1. Highway Information System (HIS)

The Highway Information System (HIS) offers a comprehensive repository of roadway data, systematically organized into eight primary categories and over 30 detailed subgroups. Each subgroup is structured using route-specific identifiers namely the Route ID (RT_ID), Beginning Mile Point (BMP), and Ending Mile Point (EMP) and contains critical attributes such as posted speed limits, average annual daily traffic (AADT), number of lanes, median types, and functional road classifications. For this study, 15 of these subgroups were selected based on their relevance to roadway geometry and traffic conditions, providing essential information for analyzing the transportation network across the state of Kentucky.

Using RT_ID, BMP, and EMP as linking keys, the selected HIS datasets were spatially merged into a single, comprehensive database that characterizes the entire state road network. To investigate the influence of roadway classification and homogeneity on the accuracy of individual-level injury severity prediction, the integrated dataset was stratified using three different classification schemes. The first scheme segmented the data into eight distinct groups based on functional class and road geometry (*23*). These groups included: Rural Two-Lane (R2L), Urban Two-Lane (U2L), Urban Interstate and Principal Arterial (UIP), Rural Interstate and Principal Arterial (RIP), Urban Multilane Divided (UMD), Urban Multilane Undivided (UMU), Rural Multilane Divided (RMD), and Rural Multilane Undivided (RMU). This classification facilitated a fine-grained analysis of injury risk across roadway types with varying design characteristics.

The second scheme employed four binary comparisons using the entire dataset in each case. These comparisons were: Urban (U2L, UMD, UMU, and UIP) versus Rural (R2L, RMD, RMU, and RIP) roads, Freeway (UIP&RIP) versus Non-Freeway (the other six types) segments, Two-Lane (R2L&U2L) versus Multi-Lane facilities (the other six types), and Divided (UMD, RMD, UIP, RIP) versus Undivided (R2l, U2L, RMU, and UMU) highways. This approach preserved full data representation while enabling the assessment of high-level differences in road environments. The third and final scheme treated the data as a unified whole without any segmentation. This baseline scenario allowed for a control comparison to evaluate model performance without the confounding effects of roadway classification.

In total, these three classification strategies yielded 17 unique datasets, eight from the detailed road groupings, eight from binary comparisons, and one from the non-segmented dataset. This robust framework supports comprehensive comparative analyses of predictive model performance across a diverse range of roadway contexts and design features.

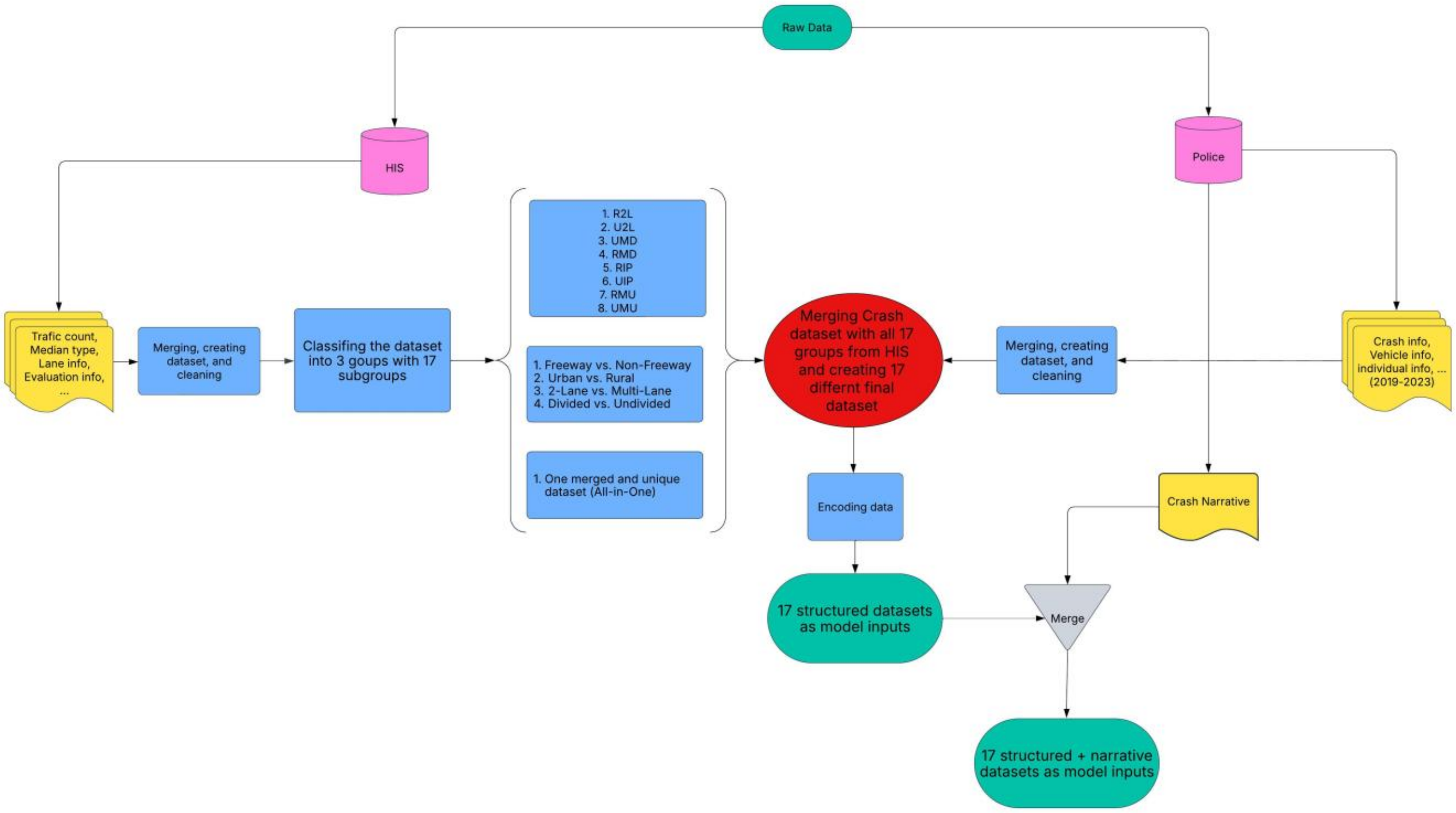

*Figure 1 Data preparation flowchart*

## 3.2. Police dataset

As illustrated in Figure 1, the second principal data source for this study is the Kentucky State Police (KSP), which maintains detailed records of all officially reported crashes across the state. This research utilizes five years of crash data spanning from 2019 to 2023. The KSP crash data is systematically categorized into six primary groups: (1) general crash information, (2) vehicle-level data, (3) person-level details, (4) environmental conditions, (5) vehicle-specific factors, and (6) driver-related characteristics. For each year, datasets representing these six categories were acquired, cleaned, and merged to create a unified person-level crash dataset. These annual datasets were then combined to generate a comprehensive five-year crash database for use in this study.

Given the individual-level focus of the research, only records corresponding to persons involved in crashes were retained. The dataset identifies four injury severity levels: Fatal, Serious Injury, Minor Injury, and Possible Injury. Unfortunately, the dataset does not contain a clear label for "No Injury" cases at the person level. To address this gap, an initial strategy was implemented in which individuals involved in crashes labeled as Property Damage Only (PDO) with no recorded injuries were treated as "No Injury" cases. However, this assumption introduced noise and degraded model performance. As a result, the final study excludes the "No Injury" class and focuses exclusively on the four recorded injury categories for predictive modeling.

Subsequently, as indicated by the red circle in Figure 1, the final cleaned crash dataset was spatially joined with all 17 datasets derived from the Highway Information System (HIS). This integration was performed using the Route ID (RT_ID), Beginning Mile Point (BMP), Ending Mile Point (EMP), and the specific Mile Point of each crash to accurately align crash records with corresponding roadway attributes. After merging, the combined dataset was carefully reviewed to resolve redundancies and conflicting values. For example, a discrepancy of approximately 2% was observed between the posted speed limits reported in HIS and those recorded by the police. Given that police records reflect real-time conditions and temporary adjustments (e.g., due to construction zones or enforcement areas), their values were considered more accurate and retained in the final dataset.

In the final step, as shown in Figure 1, each of the 17 structured datasets was merged with the crash narrative feature to generate the final 17 crash datasets. Crash narratives are unstructured, free-text descriptions written by police officers at the scene of an incident. These narratives capture detailed accounts of the crash sequence, contributing factors, driver behavior, and road or environmental conditions contextual information that is often missing from structured database fields. Integrating this narrative data provides additional depth and realism for injury severity prediction.

To enable machine learning analysis, categorical features were transformed into numerical or binary variables using appropriate encoding methods. Following this preprocessing, the finalized dataset included 92 main features and over 350 derived sub-features. A summary of key variables used for modeling is provided in Table 1.

**Table 1**
Crash percentages and features by varying crash severity.

| Feature | Category | Number | Percentage of Traffic Crash Severity (%) | | | |
|---|---|---|---|---|---|---|
| | | | Fatal | Major Injury | Minor Injury | Possible Injury |
| Speed Limit | 0 - 25 | 1110 | 1.09 | 6.84 | 40.27 | 51.80 |
| | 30 - 50 | 10503 | 1.96 | 7.40 | 39.28 | 5136 |
| | 55 - 75 | 20727 | 4.15 | 12.42 | 38.58 | 44.85 |
| Gender | Male | 16583 | 4.45 | 13.23 | 39.73 | 42.59 |
| | Female | 15713 | 2.16 | 7.85 | 38.00 | 52.00 |
| Age | 22 > | 7026 | 2.01 | 9.05 | 38.86 | 50.09 |
| | 22 - 65 | 21507 | 3.49 | 11.24 | 39.36 | 45.91 |
| | > 65 | 3412 | 5.33 | 10.11 | 38.04 | 46.51 |
| Alcohol | Yes | 1609 | 9.63 | 22.68 | 40.09 | 27.59 |
| | No | 21196 | 2.91 | 10.10 | 39.94 | 47.05 |
| Crash Type | Angle | 4762 | 2.33 | 6.22 | 36.92 | 54.54 |
| | Backing | 66 | 1.52 | 3.03 | 39.39 | 56.06 |
| | Head on | 2634 | 8.12 | 19.13 | 35.73 | 37.02 |
| | Left turn | 1020 | 0.98 | 8.14 | 44.41 | 46.47 |
| | Rear end | 7710 | 1.14 | 4.81 | 35.67 | 58.38 |
| | Rear to rear | 137 | 0.73 | 6.57 | 38.69 | 54.01 |
| | Sideswipe-O-Dir | 1064 | 3.38 | 8.74 | 42.39 | 45.49 |
| | Sideswipe-Dir | 1732 | 0.92 | 6.12 | 40.01 | 52.94 |
| | Single vehicle | 13214 | 4.55 | 14.86 | 41.19 | 39.40 |
| Light Condition | Dawn | 702 | 2.99 | 7.12 | 40.46 | 49.43 |
| | Daylight | 21891 | 2.63 | 9.68 | 37.73 | 49.96 |
| | Dusk | 746 | 5.63 | 14.75 | 41.96 | 37.67 |
| | Dark-HWY lighted/off | 491 | 4.68 | 15.07 | 39.31 | 40.94 |
| | Dark-HWY lighted/on | 3006 | 3.06 | 9.25 | 44.64 | 43.05 |
| | Dark-HWY not lighted | 4948 | 6.24 | 14.96 | 38.74 | 40.06 |
| | Dark (unknown light) | 510 | 2.16 | 10.39 | 48.04 | 39.41 |
| | Other | 15 | 0.00 | 0.00 | 26.67 | 73.33 |
| | Unknown | 28 | 17.86 | 10.71 | 35.71 | 35.71 |
| Airbag | On | 18342 | 3.48 | 10.39 | 39.65 | 46.48 |
| | Off | 814 | 2.21 | 10.32 | 40.79 | 46.68 |
| | Not Presented | 13184 | 3.2 | 10.91 | 37.65 | 48.23 |
| Secondary Crash | Yes | 705 | 4.40 | 11.91 | 39.86 | 43.83 |
| | No | 31635 | 3.31 | 10.57 | 38.84 | 47.28 |
| Weather | Blowing Snow | 36 | 2.78 | 2.78 | 69.44 | 25 |
| | Clear | 20662 | 3.58 | 11.67 | 39.28 | 45.47 |
| | Cloudy | 6192 | 3.04 | 9.9 | 36.72 | 50.34 |
| | Fog with rain | 73 | 1.37 | 6.85 | 45.21 | 46.58 |
| | Raining | 4525 | 2.63 | 7.18 | 38.43 | 51.76 |
| | Crosswinds | 45 | 0 | 17.78 | 35.56 | 46.67 |
| | Sleet, hail | 178 | 4.49 | 8.99 | 49.44 | 37.08 |
| | Snowing | 360 | 1.94 | 5.28 | 43.06 | 49.72 |
| | Fog | 205 | 5.37 | 10.24 | 43.9 | 40.49 |
| | Smog, smoke | 7 | 0 | 28.57 | 28.57 | 42.86 |
| | Other | 55 | 7.27 | 9.09 | 52.73 | 30.91 |
| Person Type | Driver | 22806 | 3.39 | 10.99 | 39.95 | 45.68 |
| | Passenger | 8664 | 2.1 | 8.75 | 35.66 | 53.49 |
| | Pedestrian | 668 | 16.62 | 20.21 | 39.37 | 23.8 |
| | Animal-drawn/ridden | 29 | 0 | 20.69 | 41.38 | 37.93 |
| | Bicyclist | 159 | 8.18 | 13.21 | 54.09 | 24.53 |
| | Property owner | 14 | 0 | 14.29 | 57.14 | 28.57 |

# 4. Methodology

This section outlines the comprehensive framework adopted to develop and evaluate injury severity prediction models using a hybrid dataset that merges structured crash data with unstructured police narrative texts. The approach includes detailed preprocessing techniques for both numerical/categorical variables and free-text crash narratives. For textual data, natural language processing (NLP) techniques—specifically Term Frequency–Inverse Document Frequency (TF-IDF) and Word2Vec (W2V), were employed to convert narrative content into numerical vector representations that capture semantic meaning and contextual relevance. The dataset was strategically partitioned into training, validation, and testing subsets, allowing for model tuning using the validation set before final evaluation. Class imbalance in injury severity outcomes was addressed using the Synthetic Minority Over-sampling Technique (SMOTE) in combination with K-Nearest Neighbors (KNN). Feature selection for the structured data was conducted using Random Forest importance scores, and the top 100 most influential features were retained to reduce dimensionality and enhance model performance. This section also details the machine learning algorithms employed—Random Forest (RF), AdaBoost, and XGBoost—and their integration with structured and narrative-derived features to build robust models capable of capturing both quantitative patterns and qualitative contextual cues from crash descriptions.

## 4.1. Data Preprocessing: Handling Missing Values

Real-world datasets, particularly those compiled from crash reports, frequently suffer from missing values. These can arise from various factors, including incomplete data entry or information not being collected for certain incidents. Effective management of missing data is crucial as it can significantly impact model training and predictive performance. In this study, a systematic two-pronged strategy was implemented:

### 4.1.1. Feature Exclusion for High Missingness

Features (columns) exhibiting a high proportion of missing values were deemed unreliable and removed from the dataset. Specifically, any feature where the percentage of missing observations exceeded 50% was discarded. This threshold, commonly adopted in data cleaning, prevents the introduction of substantial noise or bias that could result from extensively imputing highly incomplete variables (*24*). If $N_{missing_i}$ represents the number of missing values for feature i, and $N_{total}$ is the total number of observations, then feature i was removed if:

$$\frac{N_{missing_i}}{N_{total}} > 0.50$$

### 4.1.2. Imputation for Moderate Missingness

For features with less than 50% missing values, imputation techniques were applied to fill in the gaps:

#### *4.1.2.1. Categorical Features*

Missing values in categorical features were imputed using the mode, which is the most frequently occurring category. This approach is simple and effective, preserving the distribution of the existing categorical data and preventing the introduction of new, unobserved categories (*25*). For a categorical feature C, the imputed value $C_{\text{imputed}}$ for any missing entry is given by:

$$C_{\text{imputed}} = Mode(C)$$

#### *4.1.2.2. Numerical Features*

Missing values in numerical features were imputed using the median. The median is the middle value in a numerically ordered dataset. It is a robust measure of central tendency that is less sensitive to outliers or skewed distributions compared to the mean (*24; 25*). For a numerical feature F, the imputed value $F_{\text{imputed}}$ for any missing entry is given by:

$$F_{\text{imputed}} = Median(F)$$

This combination of deletion for severely incomplete features and targeted imputation for moderately incomplete ones ensures data quality while maximizing information retention.

## 4.2. Data Splitting and Validation Strategy

To ensure robust model development, prevent overfitting, and obtain an unbiased assessment of generalization performance, the entire dataset was partitioned into three distinct subsets: training, validation, and testing, an essential strategy for reliable machine learning experimentation (*26*). The training set, which typically comprises 70% of the data, was used to train the machine learning models by allowing them to learn complex patterns and relationships between input features and the target injury severity outcomes. The validation set, usually around 15% of the data, served for model selection and hyperparameter tuning. During model development, different configurations were trained on the training set and evaluated on the validation set to determine the best-performing model or optimal parameters—without introducing data leakage from the final test set (*27*). The final subset, the testing set, comprising the remaining 15%, was used only once for the final, unbiased evaluation of the selected model. This simulates real-world application by exposing the model to entirely unseen data. To preserve class distribution across these subsets, particularly given the class imbalance inherent in crash data, stratified sampling was employed during the split.

Using a three-way data split provides several methodological advantages compared to a traditional train/test split. Most importantly, it prevents data leakage by completely isolating the test set, ensuring that no performance information influences model selection or tuning (*28*). This setup leads to a more honest and accurate assessment of the model's ability to generalize new data. Additionally, the use of a separate validation set allows for efficient and iterative hyperparameter tuning, helping optimize model performance without compromising the integrity of the final evaluation. Together, this methodology enhances the credibility, reliability, and real-world applicability of the predictive model (*29*).

## 4.3. Addressing Class Imbalance: Oversampling with k-Nearest Neighbors (SMOTE)

Traffic crash datasets are characterized by severe class imbalance, where non-injury or minor injury outcomes vastly outnumber severe injury or fatal cases. Training machine learning models directly on such imbalanced data can lead to models that are heavily biased towards the majority class, resulting in suboptimal performance, particularly in detecting the critical, rare severe injury instances. To mitigate this issue, the Synthetic Minority Oversampling Technique (SMOTE) was applied exclusively to the training dataset.

SMOTE generates synthetic samples for the minority class(es) by interpolating between existing minority class instances and their nearest neighbors, rather than merely duplicating them. This approach helps to expand the minority class's representation in the training data while reducing the risk of overfitting that can occur with simple random oversampling (*30*) . The process for generating a synthetic sample is as follows: 1- For each minority class sample $x_i$, its K nearest neighbors from the same minority class are identified. 2- One of these K neighbors, say $x_j$, is randomly selected. 3- A new synthetic sample $x_{new}$ is created along the line segment connecting $x_i$ and $x_j$ using the following formula:

$$x_{new} = x_i + \delta \times (x_j - x_i)$$

where δ is a random number between 0 and 1. This linear interpolation creates a new, but related, data point in the feature space.

The Synthetic Minority Oversampling Technique (SMOTE) offers several advantages for addressing class imbalance in crash severity prediction. First, by synthetically increasing the number of minority class samples, SMOTE enables learning algorithms to better capture patterns unique to these underrepresented classes, ultimately improving metrics such as precision and recall (*31*). Unlike simple duplication, SMOTE generates new, diverse instances by interpolating between existing samples, which helps reduce overfitting and encourages the formation of smoother decision boundaries (*32*). This not only mitigates the model's tendency to memorize rare cases but also improves its ability to generalize. Additionally, training on a more balanced dataset fosters classifier robustness, enhancing its performance across all classes when tested on imbalanced, real-world data. Importantly, SMOTE was applied exclusively to the training set; both the validation and test sets were left unaltered to preserve their original class distributions. This ensures that performance evaluations remain unbiased and truly reflective of the model's effectiveness in real-world deployment scenarios (*33*).

## 4.4. Feature Selection

Feature selection is a critical preprocessing step for high-dimensional datasets. Its primary goals are to reduce model complexity, mitigate the risk of overfitting, decrease training time, and enhance model interpretability and generalization by identifying and retaining only the most relevant input variables (*34-36*). In this study, the intrinsic feature importance capabilities of the Random Forest algorithm were leveraged for this purpose. The Random Forest algorithm intrinsically provides a mechanism for quantifying the relative importance of input features. This is typically achieved by calculating the Mean Decrease in Impurity (MDI), also known as Gini Importance (*37; 38*).For a decision tree, impurity (e.g., Gini impurity for classification) at a node is a measure of the heterogeneity of the classes within that node. Gini impurity for a node with K classes is given by:

$$G(p) = 1 - \sum_{K=1}^{K} p_k^2$$

where $p_k$ is the proportion of samples belonging to class k at that node.

When a feature is used to split a node, the impurity of the child nodes is generally lower than that of the parent node. The importance of a feature is derived by summing the weighted decrease in impurity over all splits where that feature is used across all decision trees in the Random Forest ensemble. This sum is then averaged over all trees, providing a global importance score for each feature. Features that consistently lead to a greater reduction in impurity are deemed more important predictors of the target variable. The methodology for feature selection involved: 1- Extracting the MDI-based feature importance scores for all features from this trained Random Forest model. 2- Ranking all features in descending order based on their importance scores. 3- Selecting the Top 100 Features: Through an iterative empirical process, different numbers of top features (e.g., top 50, top 150, or all available features) were used to train subsequent models, and their performance was evaluated on the validation set. It was consistently found that utilizing the top 100 most important features yielded the highest overall model performance. This indicates an optimal balance where sufficient predictive power is retained while minimizing noise and computational overhead.

Random Forest is a powerful tool for feature selection, offering several advantages that make it well-suited for complex datasets. One key strength is its robustness to multicollinearity; Random Forests can handle correlated features without significantly distorting their importance measures. Moreover, they are capable of capturing non-linear relationships and interaction effects between features and the target variable—patterns that traditional linear methods often overlook (*39*).This allows Random Forests to uncover meaningful predictors even when the data structure is intricate. Another major advantage is the model-agnostic nature of its output: once important features are identified, they can be used in any subsequent machine learning model, not just Random Forests. This makes it a flexible and effective dimensionality reduction technique that enhances performance and interpretability across a range of predictive modeling tasks.

## 4.5. Machine Learning Models

This study employed three powerful ensemble machine learning algorithms that are widely recognized for their high predictive accuracy, robustness, and ability to handle complex, tabular data classification tasks: Random Forest, AdaBoost, and XGBoost. Ensemble learning methods operate on the principle of combining predictions from multiple individual "weak learners" to produce a more accurate and stable overall prediction than any single learner could achieve independently (*40*).

### 4.5.1. Random Forest (RF)

Random Forest, introduced by Breiman (2001), is a widely used and versatile ensemble learning method grounded in the principle of bagging (bootstrap aggregating). It builds a collection of decision trees during training and makes predictions by aggregating the outputs of these trees, using majority voting for classification tasks. The method relies on two key mechanisms: bootstrap sampling and random feature subspace selection. In bootstrap sampling, each decision tree is trained on a randomly drawn dataset (with replacement) from the original training data, which helps reduce model variance. Additionally, at each node split during tree construction, only a random subset of features is considered. This technique, known as the random subspace method, decorrelates the trees and enhances the ensemble's robustness to noise, leading to improved generalization performance. The trees in the forest are trained in parallel, and their predictions are aggregated to form the final output. Random Forest offers several advantages: it delivers high predictive accuracy, performs well on high-dimensional datasets, is resilient to outliers and noisy data, is less prone to overfitting than individual decision trees, and naturally provides feature importance measures, making it valuable both for modeling and feature selection tasks (*38*).

### 4.5.2. AdaBoost (Adaptive Boosting)

AdaBoost (Adaptive Boosting), one of the earliest and most influential boosting algorithms, was introduced by Freund and Schapire (1997). It works by iteratively training a sequence of weak learners, most commonly simple decision trees (often referred to as "decision stumps" if they consist of only one split). The core mechanism of AdaBoost involves: 1- Iterative Weight Adjustment: In each boosting iteration, AdaBoost adjusts the weights of the training instances. Instances that were misclassified by the previous weak learner receive higher weights for the next iteration, while correctly classified instances receive lower weights. This forces subsequent weak learners to focus more on the "difficult" samples. 2- Weighted Combination of Weak Learners: After training a specified number of weak learners, the final strong classifier H(x) is constructed as a weighted sum of the individual weak learners $h_t$(x), where each weak learner $h_t$(x) is assigned a weight αt based on its accuracy (*41*):

$$H(x) = sign(\sum_{t=1}^{T} \alpha_t h_t\,(x))$$

### 4.5.3. XGBoost (Extreme Gradient Boosting)

XGBoost (Extreme Gradient Boosting) is an optimized, distributed gradient boosting library that has gained immense popularity for its efficiency, flexibility, and high predictive performance in various machine learning competitions and real-world applications (*42*). It builds upon the general framework of gradient boosting, where new decision trees are sequentially added to correct the errors (residuals) made by previous trees. Key enhancements and features of XGBoost include: 1- Regularization: XGBoost incorporates L1 (Lasso) and L2 (Ridge) regularization terms into its objective function. This helps to prevent overfitting by penalizing the complexity of the trees, leading to more generalized models. The regularized objective function at iteration t can be conceptually represented as:

$$Obj^{(t)} = \sum_{i=1}^{n} l(y_i\,, y_i^{(t-1)} + f_t(x_i)) + \Omega(f_t)$$

where $l$ is the loss function, $y_i^{(t-1)}$ is the prediction from the previous (t−1) trees, $f_t(x_i)$ is the new tree to be added, and $\Omega(f_t)$is the regularization term for the new tree. 2- Parallel Processing: Optimized to run efficiently on parallel and distributed computing environments, significantly speeding up training times for large datasets. 3- Handling Missing Values: XGBoost can inherently handle missing values. It learns the optimal direction for missing values to go (left or right child node) during tree splits based on patterns observed in the data. 4- Tree Pruning: Unlike greedy approaches, XGBoost grows trees to their maximum depth first and then prunes them backward based on a specified gain criterion, which helps in preventing overfitting and finding optimal tree structures (*43*).

## 4.6. Textual Feature Extraction: Natural Language Processing (NLP) Techniques

Crash narratives, being free-form text, require specialized techniques to convert linguistic information into a structured, numerical format suitable for machine learning algorithms. This study utilized two distinct approaches for textual feature extraction: TF-IDF for sparse, weighted term representations and Word2Vec for dense, contextual word embeddings.

### 4.6.1. TF-IDF (Term Frequency-Inverse Document Frequency)

TF-IDF is a widely used statistical measure for evaluating the importance of a word in a document relative to a collection of documents (corpus). It assigns a weight to each term based on how frequently it appears in a specific document (Term Frequency) balanced by how rare it is across the entire corpus (Inverse Document Frequency) (*44*). The TF-IDF transformation was applied to the preprocessed crash narratives. The process involves:

#### *4.6.1.1. Term Frequency (TF)*

Measures how frequently a term t appears in a document d. This can be a raw count or a normalized count.

$$TF(t, d) = frequency\ of\ term\ t\ in\ document\ d$$

#### *4.6.1.2. Inverse Document Frequency (IDF)*

Measures how important a term is by down-weighting terms that appear very frequently across many documents (e.g., stop words) and up-weighting terms that are rare. The IDF for a term t in a corpus D with N documents and $df(t)$ being the number of documents containing term t is calculated as:

$$IDF(t, D) = \log\left(\frac{N}{1 + df(t)}\right) + 1$$

#### *4.6.1.3. TF-IDF Score*

The final TF-IDF score for a term t in document d within corpus D is the product of its TF and IDF scores, typically followed by L2 normalization for the entire document vector:

$$TF_IDF = TF(t, d) \times IDF(t, , D)$$

Each processed narrative was transformed into a sparse TF-IDF vector, representing the narrative in a high-dimensional feature space where each dimension corresponds to a unique term from the corpus vocabulary. These sparse vectors were then converted into dense NumPy arrays for compatibility with the machine learning models.

TF-IDF offers several advantages for analyzing narrative crash data. It effectively highlights keywords that are contextually important within individual narratives but relatively rare across the broader dataset, enabling the identification of critical terms that may signal specific injury scenarios or crash dynamics. Its computational efficiency makes it well-suited for large-scale datasets, allowing for rapid processing without significant resource demands. Additionally, the TF-IDF vectorization process naturally emphasizes the most informative words, providing an inherent form of feature selection by down-weighting common or non-discriminative terms across the corpus.

### 4.6.2. Word2Vec Embeddings

To overcome the limitations of sparse, bag-of-words representations like TF-IDF (which do not capture semantic relationships or word context), Word2Vec was utilized to generate dense, continuous vector representations (embeddings) of words from the crash narratives(*45*). Word2Vec learns these embeddings by analyzing the distributional context of words in the corpus: words that appear in similar contexts are assigned similar vector representations in a lower-dimensional space. The models aim to optimize an objective function, typically the log-likelihood of word co-occurrences. In other words, Word2Vec learns vector representations of words by analyzing their context (neighboring words) across a large corpus.

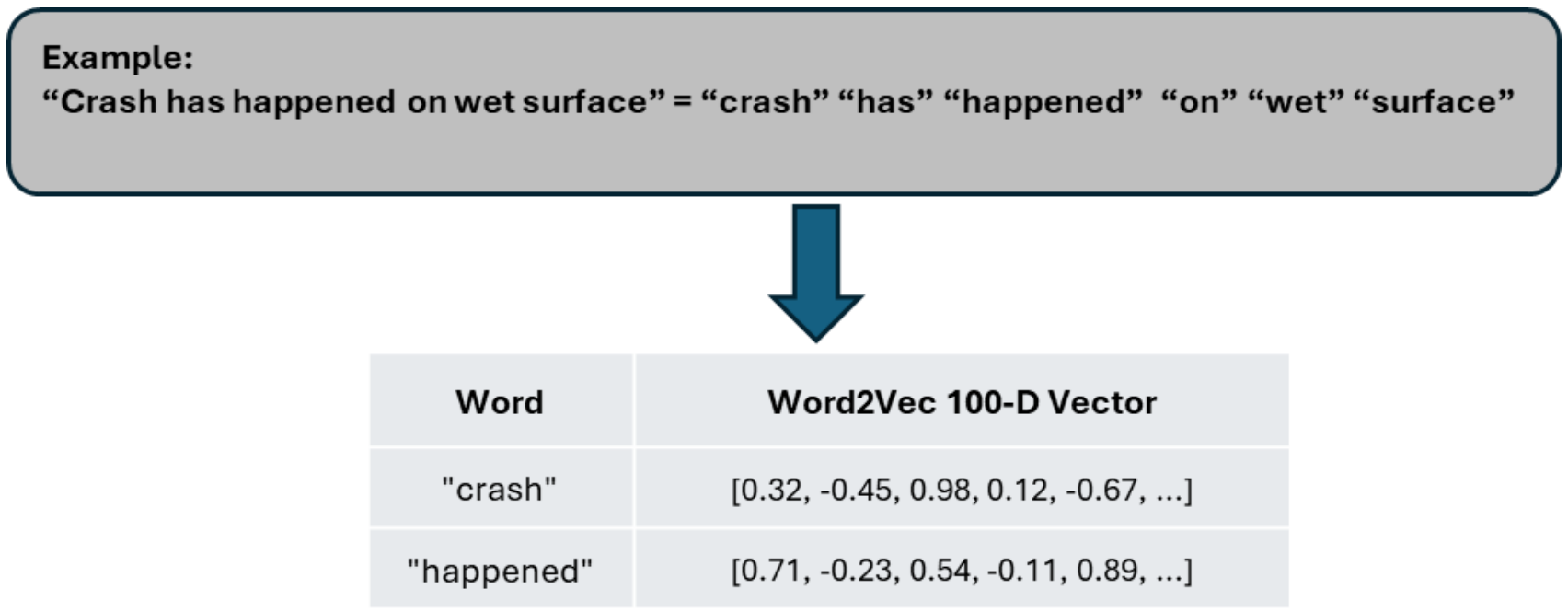

*Figure 2 simple example of embedding with W2Vector*

In this research, the vector dimension was set to 100. Although other dimensions were tested, using 100 produced more acceptable results for the predictive models. For instance, if a crash narrative contains 200 words, the resulting vector matrix will have dimensions of 200×100. See the example in figure 2.

Word2Vec embeddings offer several advantages for text representation in predictive modeling. They effectively capture both semantic and syntactic relationships, meaning that words with similar meanings (e.g., "collision" and "crash") or similar grammatical functions tend to occupy nearby positions in the embedding space. Unlike high-dimensional and sparse representations such as TF-IDF, Word2Vec produces dense, lower-dimensional vectors, which not only reduce computational complexity but also enhance the model’s ability to generalize. Additionally, Word2Vec provides a deeper contextual understanding by encoding the surrounding context of each word, allowing models to leverage rich narrative information beyond simple term frequency.

These textual features, derived from both TF-IDF and Word2Vec, were then integrated (e.g., by concatenation) with the structured data features to form a comprehensive, multi-modal input for the machine learning models, enabling a more holistic understanding of crash severity factors.

## 4.7. Evaluation Metrics

Given the confirmed class imbalance in the test dataset, model performance was rigorously evaluated using metrics that are robust to skewed class distributions. Relying solely on accuracy would be misleading in such scenarios. The primary evaluation metrics were:

### 4.7.1. Precision (P)

The proportion of correctly predicted positive observations out of all positive predictions. It quantifies the model's ability to avoid false positives.

$$P = \frac{\text{True Positives (TP)}}{\text{True Positives (TP)} + \text{False Positives (FP)}}$$

### 4.7.2. Recall (R) (Sensitivity)

The proportion of correctly predicted positive observations out of all actual positives. It quantifies the model's ability to find all positive instances.

$$R = \frac{\text{True Positives (TP)}}{\text{True Positives (TP)} + \text{False Negatives (FN)}}$$

### 4.7.3. F1-score (F1)

The harmonic mean of Precision and Recall. It provides a balanced measure, especially useful when there is an uneven class distribution, as it penalizes models that perform well on one metric but poorly on the other.

$$F1 = 2 \times \frac{\mathrm{P} \times \mathrm{R}}{\mathrm{P} + \mathrm{R}}$$

### 4.7.4. Macro Average F1-score

This metric calculates the F1-score for each class independently and then takes the unweighted arithmetic mean of these per-class F1-scores. It gives equal weight to each class, regardless of its size, thus providing a strong indicator of the model's performance on minority classes.

$$Marco\ F1 = \frac{1}{N_{classes}} \sum_{i=1}^{N_{classes}} F1_i$$

### 4.7.5. Weighted (Micro) Average F1-score

This metric also calculates per-class F1-scores but then takes an average weighted by the support (number of true instances) of each class. This provides an overall F1-score that reflects the performance on the larger classes more heavily, aligning with the real-world class distribution in the test set.

$$Weighted\ F1 = \sum_{i=1}^{N_{classes}} (F1_i \times \frac{Support_i}{Total\ Samples})$$

### 4.7.6. Accuracy

While secondary due to imbalance, overall accuracy providing a general measure of correctness across all predictions.

$$\text{Accuracy} = \frac{\text{True Positives (TP)} + \text{True Negatives (TN)}}{\text{True Positives (TP)} + \text{True Negatives (TN)} + \text{False Negatives (FN)} + \text{False Positives (FP)}}$$

Where the true positive (TP) is the number of positive cases correctly identified as such, the true negative (TN) is the number of negative cases correctly identified as such, the false negative (FN) is the number of positive cases incorrectly identified as negative, and the false positive (FP) is the number of negative cases incorrectly identified as positive. The process has been summarized as a flowchart in the following graph:

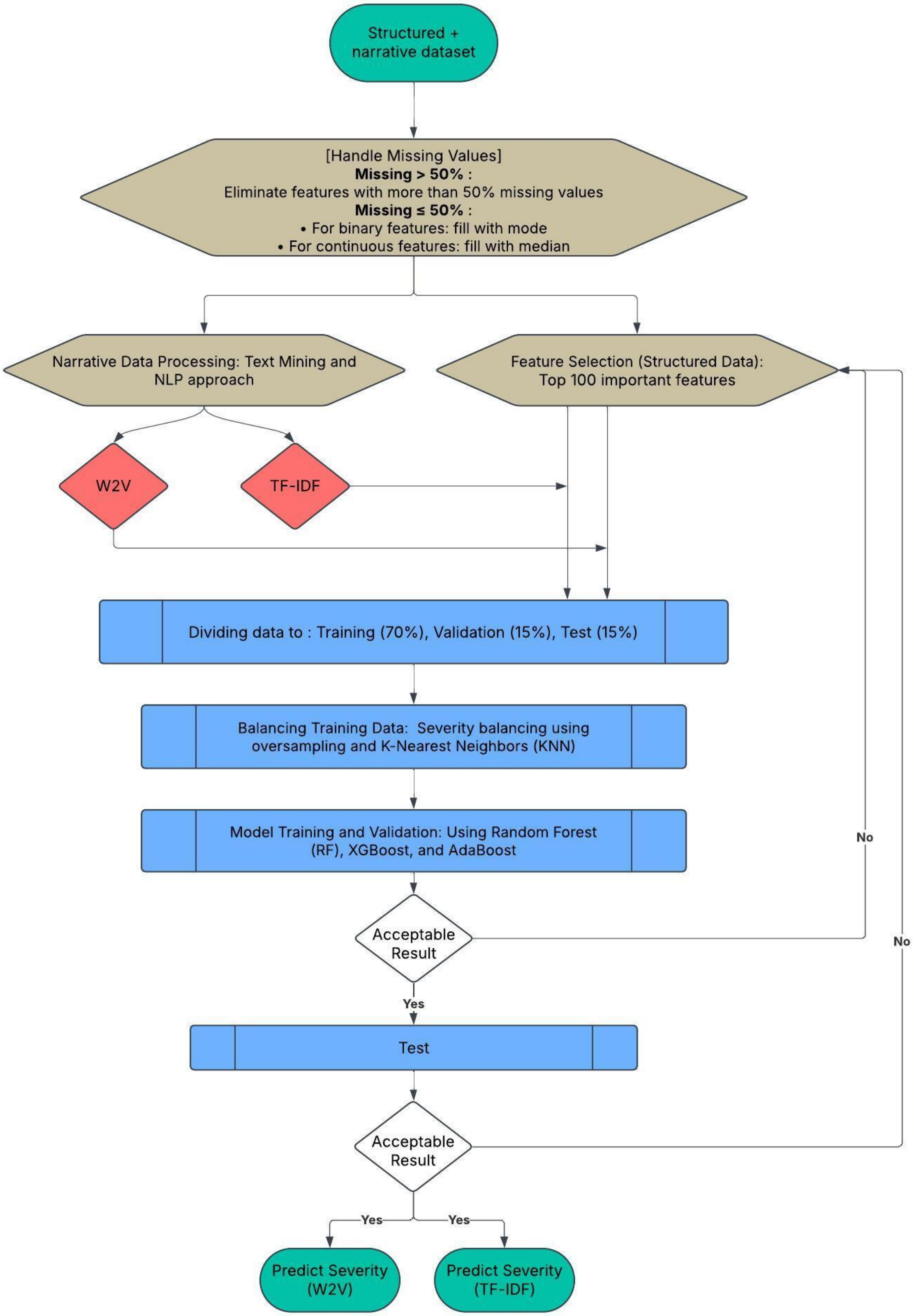

*Figure 3 Methodology flowchart*

# 5. Results and Discussion

## 5.1. Modelling summary

This section offers a detailed assessment of integrating text mining methods with machine learning algorithms to predict crash injury severity at the individual level. The evaluation spans multiple roadway classification schemes and employs standard metrics including Precision, Recall, F1-Score, and Accuracy. Notably, each results table includes a "Support" column, denoting the actual count of instances per injury category in the test set critical for interpreting the relevance and reliability of model outcomes. In total, 102 models were built, and Figures 4 and 5 highlight the top-performing models across the three learning algorithms using both text mining approaches: TF-IDF and Word2Vec. As outlined in previous sections, this study not only sought to enhance individual-level injury prediction and assess the influence of roadway categorization on model performance but also to compare the effectiveness of two widely used text representation techniques. The dataset was initially stratified into eight distinct classes (Group 1); however, due to the rarity of certain severe outcomes like fatalities, models trained on those classes often yielded unstable or misleading evaluation scores. For instance, Figures 4 and 5 show that models built for classes such as UMD and UMU exhibited either inflated or suppressed metrics, largely due to insufficient test records. To mitigate this, Group 2 adopted a more aggregated classification by combining the original categories into four broader groupings, which improved sample sizes and led to more dependable model outcomes. Lastly, Group 3 treated the entire dataset as a unified set without any roadway-type division, allowing for a holistic benchmark to compare against the segmented modeling strategies employed in Groups 1 and 2.

In this study, road classification in Group 1 served as the baseline framework for comparative model evaluation. For each road class, the best-performing model was identified through a comprehensive assessment of multiple performance metrics. Given that the test sets were deliberately left imbalanced to maintain the natural distribution of injury severity levels, overall accuracy was not relied upon as the primary performance measure. Instead, the macro-averaged F1-score was used as the principal metric, as it assigns equal importance to all severity classes regardless of their prevalence. This choice is particularly important in multi-class, imbalanced classification problems, where metrics such as accuracy or micro-average F1-score may be biased toward majority classes, potentially masking poor performance on rare but critical categories such as fatal injuries. Supplementary evaluation involved examining macro-averaged precision and recall, along with class-specific metrics, to verify that the model performed reasonably well across all severity types. Although micro and weighted averages can still provide insights in some applications, they disproportionately reflect the dominant classes and are less informative in assessing model robustness on underrepresented categories. Therefore, macro-averaged metrics offer a more equitable and interpretable basis for evaluating predictive performance in this context.

All model results and evaluation metrics are presented in Figures 4 and 5. For instance, the ranking process for identifying the top three models for the Urban Two-Lane (U2L) category under the combination of machine learning with TF-IDF begins by examining the macro-average F1-score for Group 1 and its U2L subgroup, which is 0.630. Next, the evaluation proceeds to Group 2. Since U2L falls under the “Urban” classification in the first pair of Group 2, the corresponding macro-average F1-score is 0.634. In the next subgroup of Group 2, U2L is included under the “Undivided” category, where the score is 0.628. When considering the “Two-Lane” category within Group 2, the score is 0.611, and for the “Non-Freeway” category the score is 0.615. Finally, the macro-average F1-score for Group 3 is 0.644. Based on these results, the highest-performing configuration for U2L corresponds to Group 3, followed by Group 2 (Urban), and then Group 1. The same ranking procedure was applied to the other roadway classes.

## Group 1

| Word2Vec | | | | | | | |
|---|---|---|---|---|---|---|---|
| Road Class | Model | Injury type | Precision | Recall | F1-score | Support | Accuracy |
| R2L | XG | Fatal | 0.774 | 0.793 | 0.783 | 82 | 0.594 |
| | | Major injury | 0.484 | 0.446 | 0.464 | 267 | |
| | | Minor injury | 0.562 | 0.545 | 0.553 | 651 | |
| | | Possible injury | 0.638 | 0.673 | 0.655 | 688 | |
| | | Macro Avg | 0.614 | 0.614 | 0.614 | 1688 | |
| | | Micro Avg | 0.591 | 0.594 | 0.592 | | |
| RIP | RF | Fatal | 0.667 | 0.743 | 0.703 | 35 | 0.595 |
| | | Major injury | 0.425 | 0.388 | 0.405 | 80 | |
| | | Minor injury | 0.532 | 0.558 | 0.545 | 283 | |
| | | Possible injury | 0.674 | 0.654 | 0.664 | 373 | |
| | | Macro Avg | 0.574 | 0.586 | 0.579 | 771 | |
| | | Micro Avg | 0.596 | 0.595 | 0.595 | | |
| RMD | there is not enough number of records for K and A types to train and test (includes just one A and zero K) | | | | | | |
| RMU | there is not enough number of records for K type to train and test (includes just one K) | | | | | | |
| U2L | XG | Fatal | 0.824 | 0.875 | 0.848 | 16 | 0.625 |
| | | Major injury | 0.490 | 0.333 | 0.397 | 75 | |
| | | Minor injury | 0.574 | 0.638 | 0.605 | 351 | |
| | | Possible injury | 0.685 | 0.658 | 0.671 | 403 | |
| | | Macro Avg | 0.643 | 0.626 | 0.630 | 845 | |
| | | Micro Avg | 0.624 | 0.625 | 0.622 | | |
| UIP | XG | Fatal | 0.567 | 0.739 | 0.642 | 23 | 0.622 |
| | | Major injury | 0.575 | 0.354 | 0.438 | 65 | |
| | | Minor injury | 0.548 | 0.489 | 0.517 | 448 | |
| | | Possible injury | 0.672 | 0.744 | 0.706 | 616 | |
| | | Macro Avg | 0.590 | 0.581 | 0.575 | 1152 | |
| | | Micro Avg | 0.616 | 0.622 | 0.616 | | |
| UMD | XG | Fatal | 0.500 | 1.000 | 0.667 | 1 | 0.584 |
| | | Major injury | 0.364 | 0.333 | 0.348 | 12 | |
| | | Minor injury | 0.486 | 0.500 | 0.493 | 70 | |
| | | Possible injury | 0.680 | 0.667 | 0.673 | 102 | |
| | | Macro Avg | 0.507 | 0.625 | 0.545 | 185 | |
| | | Micro Avg | 0.585 | 0.584 | 0.584 | | |
| UMU | RF | Fatal | 0.600 | 1.000 | 0.750 | 3 | 0.639 |
| | | Major injury | 0.545 | 0.429 | 0.480 | 14 | |
| | | Minor injury | 0.577 | 0.562 | 0.569 | 73 | |
| | | Possible injury | 0.692 | 0.713 | 0.702 | 101 | |
| | | Macro Avg | 0.604 | 0.676 | 0.625 | 191 | |
| | | Micro Avg | 0.636 | 0.639 | 0.636 | | |

## Group 2

**Narrative included with W2V**

| Word2Vec | | | | | | | |
|---|---|---|---|---|---|---|---|
| Road Class | Model | Injury type | Precision | Recall | F1-score | Support | Accuracy |
| Rural | XG | Fatal | 0.838 | 0.831 | 0.834 | 118 | 0.578 |
| | | Major injury | 0.469 | 0.453 | 0.461 | 349 | |
| | | Minor injury | 0.528 | 0.544 | 0.536 | 944 | |
| | | Possible injury | 0.631 | 0.621 | 0.626 | 1069 | |
| | | Macro Avg | 0.616 | 0.612 | 0.614 | 2480 | |
| | | Micro Avg | 0.579 | 0.578 | 0.578 | | |
| Urban | XG | Fatal | 0.725 | 0.841 | 0.779 | 44 | 0.643 |
| | | Major injury | 0.631 | 0.424 | 0.507 | 165 | |
| | | Minor injury | 0.579 | 0.535 | 0.556 | 942 | |
| | | Possible injury | 0.683 | 0.749 | 0.715 | 1221 | |
| | | Macro Avg | 0.655 | 0.637 | 0.639 | 2372 | |
| | | Micro Avg | 0.639 | 0.643 | 0.639 | | |

| Word2Vec | | | | | | | |
|---|---|---|---|---|---|---|---|
| Road Class | Model | Injury type | Precision | Recall | F1-score | Support | Accuracy |
| Divided | XG | Fatal | 0.710 | 0.817 | 0.760 | 60 | 0.638 |
| | | Major injury | 0.504 | 0.380 | 0.433 | 158 | |
| | | Minor injury | 0.588 | 0.527 | 0.556 | 806 | |
| | | Possible injury | 0.678 | 0.748 | 0.711 | 1095 | |
| | | Macro Avg | 0.620 | 0.618 | 0.615 | 2119 | |
| | | Micro Avg | 0.632 | 0.639 | 0.633 | | |
| Undivided | XG | Fatal | 0.808 | 0.784 | 0.796 | 102 | 0.595 |
| | | Major injury | 0.507 | 0.407 | 0.452 | 356 | |
| | | Minor injury | 0.554 | 0.575 | 0.565 | 1080 | |
| | | Possible injury | 0.635 | 0.653 | 0.644 | 1195 | |
| | | Macro Avg | 0.626 | 0.605 | 0.614 | 2733 | |
| | | Micro Avg | 0.593 | 0.595 | 0.593 | | |

| Word2Vec | | | | | | | |
|---|---|---|---|---|---|---|---|
| Road Class | Model | Injury type | Precision | Recall | F1-score | Support | Accuracy |
| Multi Lanes | XG | Fatal | 0.743 | 0.825 | 0.782 | 63 | 0.642 |
| | | Major injury | 0.483 | 0.399 | 0.437 | 173 | |
| | | Minor injury | 0.582 | 0.528 | 0.554 | 884 | |
| | | Possible injury | 0.691 | 0.751 | 0.720 | 1200 | |
| | | Macro Avg | 0.624 | 0.626 | 0.623 | 2320 | |
| | | Micro Avg | 0.635 | 0.642 | 0.637 | | |
| 2 Lanes | XG | Fatal | 0.776 | 0.768 | 0.772 | 99 | 0.581 |
| | | Major injury | 0.465 | 0.387 | 0.422 | 341 | |
| | | Minor injury | 0.546 | 0.548 | 0.547 | 1002 | |
| | | Possible injury | 0.625 | 0.656 | 0.640 | 1090 | |
| | | Macro Avg | 0.603 | 0.590 | 0.595 | 2532 | |
| | | Micro Avg | 0.578 | 0.581 | 0.579 | | |

| Word2Vec | | | | | | | |
|---|---|---|---|---|---|---|---|
| Road Class | Model | Injury type | Precision | Recall | F1-score | Support | Accuracy |
| Freeway | XG | Fatal | 0.839 | 0.897 | 0.867 | 58 | 0.625 |
| | | Major injury | 0.466 | 0.372 | 0.414 | 145 | |
| | | Minor injury | 0.556 | 0.517 | 0.536 | 731 | |
| | | Possible injury | 0.674 | 0.726 | 0.699 | 989 | |
| | | Macro Avg | 0.634 | 0.628 | 0.629 | 1923 | |
| | | Micro Avg | 0.618 | 0.625 | 0.621 | | |
| Non-Freeway | XG | Fatal | 0.788 | 0.750 | 0.768 | 104 | 0.593 |
| | | Major injury | 0.484 | 0.420 | 0.450 | 369 | |
| | | Minor injury | 0.553 | 0.568 | 0.560 | 1155 | |
| | | Possible injury | 0.642 | 0.653 | 0.647 | 1301 | |
| | | Macro Avg | 0.617 | 0.598 | 0.606 | 2929 | |
| | | Micro Avg | 0.592 | 0.593 | 0.592 | | |

## Group 3

| Word2Vec | | | | | | | |
|---|---|---|---|---|---|---|---|
| Road Class | Model | Injury type | Precision | Recall | F1-score | Support | Accuracy |
| All together | XG | Fatal | 0.801 | 0.895 | 0.845 | 162 | 0.608 |
| | | Major injury | 0.510 | 0.414 | 0.457 | 514 | |
| | | Minor injury | 0.551 | 0.549 | 0.550 | 1885 | |
| | | Possible injury | 0.656 | 0.681 | 0.668 | 2290 | |
| | | Macro Avg | 0.629 | 0.635 | 0.630 | 4851 | |
| | | Micro Avg | 0.605 | 0.608 | 0.606 | | |

*Figure 4 Best model performance for each group and class_ Word2Vec*

Group 1

TF-IDF

| Road Class | Model | Injury type | Precision | Recall | F1-score | Support | Accuracy |
|---|---|---|---|---|---|---|---|
| R2L | XG | Fatal | 0.798 | 0.817 | 0.807 | 82 | 0.602 |
| | | Major injury | 0.526 | 0.491 | 0.508 | 267 | |
| | | Minor injury | 0.560 | 0.590 | 0.574 | 651 | |
| | | Possible injury | 0.649 | 0.631 | 0.640 | 688 | |
| | | Macro Avg | 0.633 | 0.632 | 0.632 | 1688 | |
| | | Micro Avg | 0.602 | 0.602 | 0.602 | | |
| RIP | XG | Fatal | 0.718 | 0.800 | 0.757 | 35 | 0.610 |
| | | Major injury | 0.475 | 0.363 | 0.411 | 80 | |
| | | Minor injury | 0.551 | 0.534 | 0.542 | 283 | |
| | | Possible injury | 0.660 | 0.702 | 0.681 | 373 | |
| | | Macro Avg | 0.601 | 0.600 | 0.598 | 771 | |
| | | Micro Avg | 0.603 | 0.610 | 0.605 | | |
| RMD | there is not enough number of records for K type to train and test (includes just one A and zero K) | | | | | | |
| RMU | there is not enough number of records for K type to train and test (includes just one K) | | | | | | |
| U2L | RF | Fatal | 0.857 | 0.750 | 0.800 | 16 | 0.632 |
| | | Major injury | 0.462 | 0.400 | 0.429 | 75 | |
| | | Minor injury | 0.584 | 0.632 | 0.607 | 351 | |
| | | Possible injury | 0.699 | 0.670 | 0.684 | 403 | |
| | | Macro Avg | 0.651 | 0.613 | 0.630 | 845 | |
| | | Micro Avg | 0.633 | 0.632 | 0.632 | | |
| UIP | XG | Fatal | 0.621 | 0.783 | 0.692 | 23 | 0.643 |
| | | Major injury | 0.600 | 0.369 | 0.457 | 65 | |
| | | Minor injury | 0.567 | 0.545 | 0.556 | 448 | |
| | | Possible injury | 0.697 | 0.739 | 0.717 | 616 | |
| | | Macro Avg | 0.621 | 0.609 | 0.606 | 1152 | |
| | | Micro Avg | 0.640 | 0.643 | 0.639 | | |
| UMD | XG | Fatal | 1.000 | 1.000 | 1.000 | 1 | 0.659 |
| | | Major injury | 0.286 | 0.167 | 0.211 | 12 | |
| | | Minor injury | 0.603 | 0.586 | 0.594 | 70 | |
| | | Possible injury | 0.716 | 0.765 | 0.739 | 102 | |
| | | Macro Avg | 0.651 | 0.629 | 0.636 | 185 | |
| | | Micro Avg | 0.647 | 0.659 | 0.652 | | |
| UMU | XG | Fatal | 0.500 | 1.000 | 0.667 | 3 | 0.660 |
| | | Major injury | 0.800 | 0.286 | 0.421 | 14 | |
| | | Minor injury | 0.583 | 0.575 | 0.579 | 73 | |
| | | Possible injury | 0.713 | 0.762 | 0.737 | 101 | |
| | | Macro Avg | 0.649 | 0.656 | 0.601 | 191 | |
| | | Micro Avg | 0.666 | 0.660 | 0.652 | | |

Group 2

Narrative included with TF-IDF

TF-IDF

| Road Class | Model | Injury type | Precision | Recall | F1-score | Support | Accuracy |
|---|---|---|---|---|---|---|---|
| Rural | XG | Fatal | 0.867 | 0.831 | 0.848 | 118 | 0.588 |
| | | Major injury | 0.460 | 0.433 | 0.446 | 349 | |
| | | Minor injury | 0.547 | 0.541 | 0.544 | 944 | |
| | | Possible injury | 0.633 | 0.654 | 0.643 | 1069 | |
| | | Macro Avg | 0.627 | 0.615 | 0.620 | 2480 | |
| | | Micro Avg | 0.587 | 0.588 | 0.587 | | |
| Urban | XG | Fatal | 0.712 | 0.841 | 0.771 | 44 | 0.650 |
| | | Major injury | 0.553 | 0.412 | 0.472 | 165 | |
| | | Minor injury | 0.592 | 0.558 | 0.575 | 942 | |
| | | Possible injury | 0.695 | 0.744 | 0.719 | 1221 | |
| | | Macro Avg | 0.638 | 0.639 | 0.634 | 2372 | |
| | | Micro Avg | 0.644 | 0.649 | 0.645 | | |

TF-IDF

| Road Class | Model | Injury type | Precision | Recall | F1-score | Support | Accuracy |
|---|---|---|---|---|---|---|---|
| Divided | XG | Fatal | 0.686 | 0.800 | 0.738 | 60 | 0.640 |
| | | Major injury | 0.487 | 0.367 | 0.419 | 158 | |
| | | Minor injury | 0.585 | 0.567 | 0.576 | 806 | |
| | | Possible injury | 0.690 | 0.724 | 0.707 | 1095 | |
| | | Macro Avg | 0.612 | 0.615 | 0.610 | 2119 | |
| | | Micro Avg | 0.635 | 0.640 | 0.636 | | |
| Undivided | XG | Fatal | 0.804 | 0.765 | 0.784 | 102 | 0.608 |
| | | Major injury | 0.556 | 0.461 | 0.504 | 356 | |
| | | Minor injury | 0.566 | 0.561 | 0.563 | 1080 | |
| | | Possible injury | 0.641 | 0.681 | 0.660 | 1195 | |
| | | Macro Avg | 0.642 | 0.617 | 0.628 | 2733 | |
| | | Micro Avg | 0.606 | 0.608 | 0.606 | | |

TF-IDF

| Road Class | Model | Injury type | Precision | Recall | F1-score | Support | Accuracy |
|---|---|---|---|---|---|---|---|
| Multi Lanes | XG | Fatal | 0.765 | 0.825 | 0.794 | 63 | 0.643 |
| | | Major injury | 0.515 | 0.393 | 0.446 | 173 | |
| | | Minor injury | 0.587 | 0.526 | 0.555 | 884 | |
| | | Possible injury | 0.682 | 0.755 | 0.717 | 1200 | |
| | | Macro Avg | 0.637 | 0.625 | 0.628 | 2320 | |
| | | Micro Avg | 0.636 | 0.643 | 0.637 | | |
| 2 Lanes | XG | Fatal | 0.773 | 0.758 | 0.765 | 99 | 0.604 |
| | | Major injury | 0.500 | 0.405 | 0.447 | 341 | |
| | | Minor injury | 0.570 | 0.565 | 0.567 | 1002 | |
| | | Possible injury | 0.643 | 0.688 | 0.665 | 1090 | |
| | | Macro Avg | 0.622 | 0.604 | 0.611 | 2532 | |
| | | Micro Avg | 0.600 | 0.604 | 0.601 | | |

TF-IDF

| Road Class | Model | Injury type | Precision | Recall | F1-score | Support | Accuracy |
|---|---|---|---|---|---|---|---|
| Freeway | XG | Fatal | 0.839 | 0.897 | 0.867 | 58 | 0.653 |
| | | Major injury | 0.495 | 0.352 | 0.411 | 145 | |
| | | Minor injury | 0.592 | 0.558 | 0.575 | 731 | |
| | | Possible injury | 0.697 | 0.753 | 0.724 | 989 | |
| | | Macro Avg | 0.656 | 0.640 | 0.644 | 1923 | |
| | | Micro Avg | 0.646 | 0.653 | 0.648 | | |
| Non-Freeway | XG | Fatal | 0.777 | 0.769 | 0.773 | 104 | 0.600 |
| | | Major injury | 0.532 | 0.431 | 0.476 | 369 | |
| | | Minor injury | 0.551 | 0.545 | 0.548 | 1155 | |
| | | Possible injury | 0.644 | 0.684 | 0.663 | 1301 | |
| | | Macro Avg | 0.626 | 0.607 | 0.615 | 2929 | |
| | | Micro Avg | 0.598 | 0.601 | 0.598 | | |

Group 3

TF-IDF

| Road Class | Model | Injury type | Precision | Recall | F1-score | Support | Accuracy |
|---|---|---|---|---|---|---|---|
| All together | XG | Fatal | 0.793 | 0.877 | 0.833 | 162 | 0.627 |
| | | Major injury | 0.565 | 0.432 | 0.490 | 514 | |
| | | Minor injury | 0.567 | 0.559 | 0.563 | 1885 | |
| | | Possible injury | 0.671 | 0.709 | 0.689 | 2290 | |
| | | Macro Avg | 0.649 | 0.644 | 0.644 | 4851 | |
| | | Micro Avg | 0.623 | 0.627 | 0.624 | | |

*Figure 5 Best model performance for each group and class_ TF-IDF*

According to Figure 4 and 5 the best model performance from rank 1 to 3 has been summarized in Table 2 and 3:

**Table 2**
Word2VecResults for best model performance from rank 1 to 3.

| Road type | Rank | | |
|---|---|---|---|
| | Rank 1 | Rank 2 | Rank 3 |
| R2L | Group 3 | Group 2 & Undivided | Group 1 |
| RIP | Group 2 & Freeway | Group 3 | Group 2 & Multilane |
| RMD | Group 3 | Group 2 & Multilane | Group 2 & Divided |
| RMU | Group 3 | Group 2 & Multilane | Group 2 & Undivided |
| U2L | Group 2 & Urban | Group 1 | Group 3 |
| UIP | Group 2 & Urban | Group 2 & Freeway | Group 3 |
| UMD | Group 2 & Urban | Group 3 | Group 2 & Multilane |
| UMU | Group 2 & Urban | Group 3 | Group 2 & Multilane |

**Table 3**
TF-IDF Results for best model performance from rank 1 to 3.

| Road type | Rank | | |
|---|---|---|---|
| | Rank 1 | Rank 2 | Rank 3 |
| R2L | Group 3 | Group 1 | Group 2 & Undivided |
| RIP | Group 2 & Freeway | Group 3 | Group 2 & Multilane |
| RMD | Group 3 | Group 2 & Multilane | Group 2 & Rural |
| RMU | Group 3 | Group 2 & Multilane | Group 2 & Undivided |
| U2L | Group 3 | Group 2 & Urban | Group 1 |
| UIP | Group 2 & Freeway | Group 3 | Group 2 & Urban |
| UMD | Group 3 | Group 2 & Urban | Group 2 & Multilane |
| UMU | Group 3 | Group 2 & Urban | Group 2 & Multilane |

As shown in Tables 4 and 5, except for Group 2 within the Urban class, the TF-IDF text mining approach consistently outperformed Word2Vec across the other road classes and subgroups in our dataset. Among the models, XGBoost exhibited the most reliable and robust performance when applied to the combined structured and narrative data, highlighting its capability to capture complex patterns across heterogeneous inputs. Additionally, the variation in model rankings across different group classifications emphasizes the value of exploring multiple classification strategies. This flexible framework enabled us to identify the most suitable classification scheme for each road type, enhancing the accuracy of individual-level injury severity predictions. The strength of this comprehensive comparison lies in its practical applicability by tailoring the NLP technique, grouping strategy, and machine learning algorithm to each specific road context, practitioners can develop more precise and context-sensitive models for crash severity prediction.

## 5.2. Validating model-predicted crash severity using clinical injury

This section presents analyses aimed at evaluating how well the crash severity classifications predicted by the XGBoost model from group 3 correspond to clinical injury outcomes, as measured by the Injury Severity Score (ISS) derived from linked trauma registry data. As a standardized and objective measure of injury severity, ISS serves as a robust benchmark for validating model-based severity predictions derived from crash-level data.

In the first analysis, injury outcomes were categorized into four standard ISS severity classes based on clinical literature(*46*):

- Very Severe (ISS ≥ 25)
- Severe (ISS 16–24)
- Moderate (ISS 9–15)
- Minor (ISS 1–8)

These ISS classes were cross tabulated with the four ordinal crash severity levels predicted by the XGBoost model: Fatal (0), Major Injury (1), Minor Injury (2), and Possible Injury (3). Table 4 shows the distribution of ISS classifications within each model-predicted severity level.

**Table 4**
Cross-tabulation of ISS classification by XGBoost-predicted crash severity (2021-2022).

| Source of severity classification | Crash severity level | ISS classification | | | |
|---|---|---|---|---|---|
| | | Very severe (ISS: ≥25) | Severe (ISS: 16-24) | Moderate (ISS: 9-15) | Minor (ISS: 1-8) |
| XGBoost predicted severity | Fatal | 36 | 17 | 16 | 8 |
| | Major injury | 79 | 126 | 141 | 73 |
| | Minor injury | 28 | 70 | 144 | 144 |
| | Possible injury | 16 | 34 | 88 | 93 |

The results show a clear and expected pattern: higher predicted crash severity levels correspond to higher ISS classifications. For instance, a large proportion of Fatal and Major Injury predictions are associated with Very Severe and Severe ISS scores. Conversely, Possible Injury predictions predominantly fall within the Minor ISS category. Although some overlap exists between adjacent categories, especially between Major and Minor Injuries this reflects natural clinical variability and the inherent limitations of crash-level features. Nevertheless, the overall trend indicates that the XGBoost model is capable of distinguishing severity levels in a manner consistent with clinically observed injury severity.

To further validate the model's predictive alignment with clinical outcomes, a second analysis was conducted using raw (continuous) ISS values, rather than categorical groupings. This allows for a more detailed examination of injury severity across model-predicted classifications. Table 5 presents the total number of crash records within each severity level predicted by the XGBoost model, the number and percentage with available ISS data, and the corresponding average ISS values.

**Table 5**
Comparison of ISS availability across XGBoost predicted severity levels (2021-2022).

| Source | Severity level | Total records | Records with ISS available | Proportion with ISS available | Average ISS |
|---|---|---|---|---|---|
| XGBoost prediction | Fatal | 798 | 77 | 0.0965 | 23.0 |
| | Major injury | 1283 | 419 | 0.327 | 17.0 |
| | Minor injury | 8863 | 386 | 0.0436 | 11.7 |
| | Possible injury | 15856 | 231 | 0.0146 | 10.6 |

ISS availability was notably higher in more severe model-predicted categories, which is expected due to increased likelihood of hospitalization and subsequent trauma registry inclusion in those cases. Importantly, the average ISS values declined consistently from Fatal to Possible Injury categories, demonstrating a clear gradient that mirrors the model's severity scale. This trend supports the model's external validity and confirms that predicted severity levels align with actual clinical injury severity.

The distribution of raw ISS values across model-predicted severity levels is further visualized in Figure 6. The distributions are skewed toward higher ISS values in the Fatal and Major Injury groups, and more narrowly concentrated at lower ISS values in the Minor and Possible Injury groups. While there is some overlap particularly between adjacent categories such as Major and Minor Injury, this is anticipated due to the variability of clinical outcomes even among crashes with similar observable characteristics.

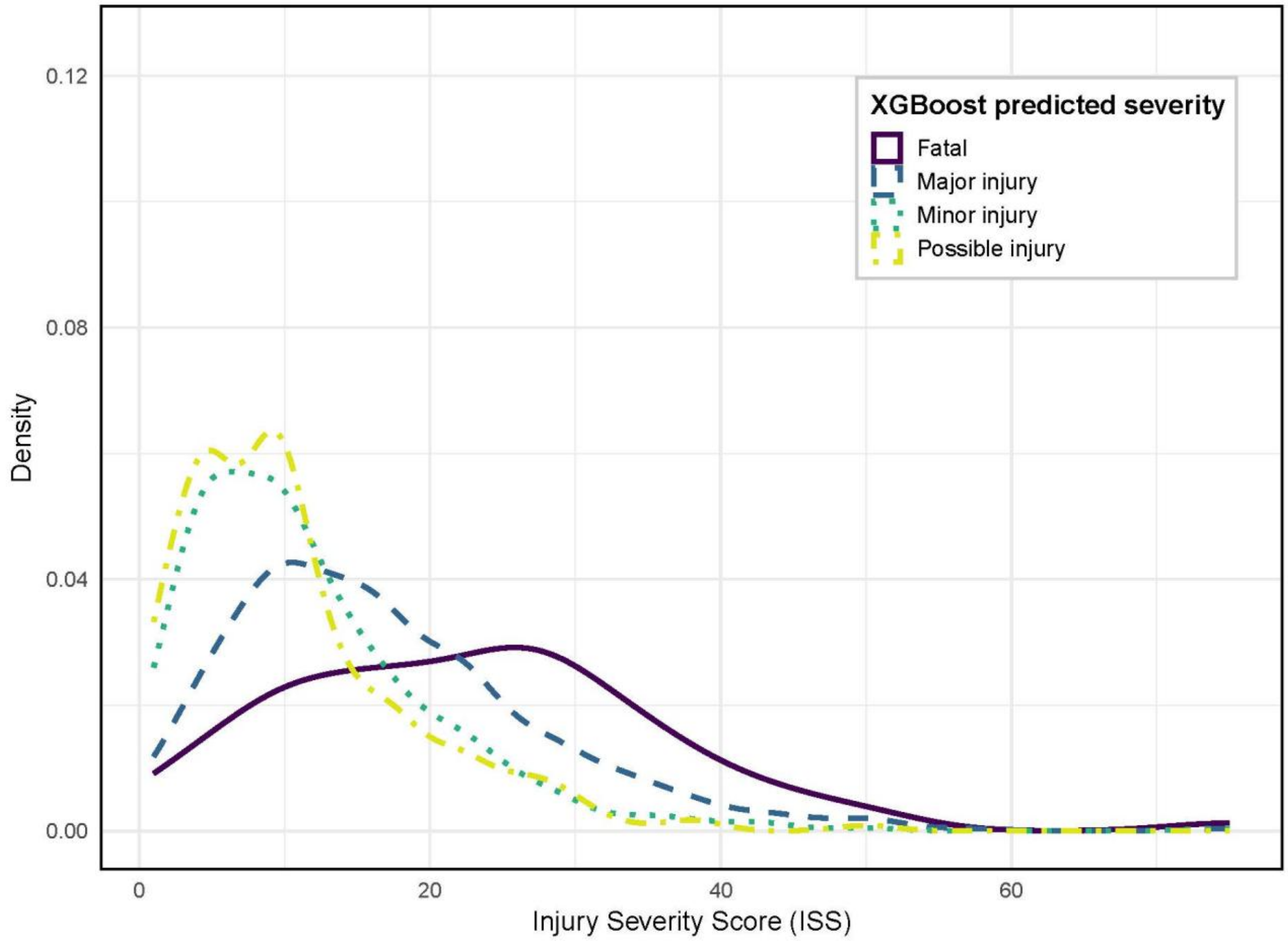

Figure 6. ISS distribution by XGBoost predicted severity.

Overall, these findings indicate a strong alignment between machine learning–predicted severity levels and real-world clinical outcomes. The consistent decline in both average ISS and distribution patterns across predicted severity levels demonstrates that the XGBoost model effectively captures meaningful distinctions in injury severity. While some overlap between intermediate categories remains reflecting the complex nature of injury mechanisms, the model's ability to stratify cases in accordance with ISS outcomes highlights its potential utility for injury severity estimation, triage prioritization, and outcome forecasting in the absence of direct clinical data.

# 6. Conclusion

This study demonstrated that integrating crash narratives with structured data markedly improves person-level injury-severity prediction, confirming that Natural Language Processing (NLP) adds valuable contextual information that conventional variables alone cannot capture. Among the three roadway-classification schemes, using different levels of segmentation provided a flexible framework for tailoring prediction strategies. Group 1 (eight detailed classes) served as a robust baseline, while Group 2's four paired categories (e.g., Freeway vs. Non-Freeway) supplied larger sample sizes for rare outcomes, yielding notably stronger results for classes such as Freeway, Multilane, and Urban. Group 3 (no classification) offered a holistic benchmark but generally ranked behind Group 2.

Across 102 narrative-enhanced models, TF-IDF was the superior text-representation technique in nearly every class except Urban in Group 2, where Word2Vec performed marginally better. XGBoost (XG) consistently delivered the highest macro F1-scores and accuracy, outperforming Random Forest and AdaBoost when handling the combined structured–narrative feature set. These findings confirm XGBoost's capacity to model complex, heterogeneous crash data.

From a practical standpoint, the framework enables timely, detailed severity prediction months before official crash data are released (e.g., 2024 data are typically unavailable until late 2025). Transportation agencies can therefore base countermeasures and resource allocation on the latest predicted injury profiles rather than delayed official statistics. Moreover, the hybrid model can impute missing severities: in our 67000-record dataset, only ~3200 cases had reported severities; the proposed methodology can fill these gaps, greatly improving data completeness for policy analysis. A limitation of the current Kentucky data is the absence of a well-defined "No-Injury" class, restricting the study to four injury levels. Incorporating this missing class in future datasets would further enhance model fidelity.

Future research can extend this work in several impactful directions. First, scenario-based analyses should be conducted by modifying key features such as, posted speed limits or work-zone presence to evaluate how these changes influence the predicted distribution of injury severities, helping identify effective policy levers. Second, the model's generalizability should be tested on crash datasets from other U.S. states or international regions to assess the robustness and transferability of the findings across diverse transportation systems. Third, more advanced text-mining approaches such as, transformer-based embeddings or topic modeling can be applied to crash narratives to enhance semantic understanding and uncover deeper causal patterns. Fourth, integrating the International Road Assessment Program (iRAP) data, such as roadway star ratings and risk factors, may enhance the model's explanatory power and provide standardized risk-context information. Fifth, incorporating Emergency Medical Services (EMS) data such as, dispatch times and on-scene response metrics could help determine whether emergency response factors play a predictive role in injury outcomes and inform EMS planning. Sixth, when reliable "No-Injury" labels become available, future models should include this additional class to provide a more complete and realistic injury-severity prediction framework. Together, these directions aim to improve model accuracy, expand applicability, and support more data-driven, life-saving transportation safety decisions.